\documentclass{simulacrum}
\usepackage{auxiliary/arxivpackages}
\usepackage{auxiliary/mymacros}
\usepackage{auxiliary/filecommands}

\renewcommand*{\backrefalt}[4]{%
    \ifcase #1 \footnotesize{(Not cited.)}%
    \or        \footnotesize{(Cited on page~#2.)}%
    \else      \footnotesize{(Cited on pages~#2.)}%
    \fi}

\renewcommand{\cite}[1]{\citep{#1}}
\setcitestyle{authoryear,square,citesep={;},aysep={,},yysep={;}}

\newcommand{\mytitle}[1][]{Chronax: A Jax Library for Forecasting and Conformal Inference}
\title[Chronax]{\mytitle}
\author{Xan Carey\affil{*} \and Arsh Singh\affil{*} \and Yash Deshmukh\affil{*} \and Sunit Jadhav\affil{*} \and Omkar Tekawade\affil{*} \and Aileen Huang\affil{*} \and Lorraine Yang\affil{*} \and Anvesha Tiwary \and Gerardo Riano \and Amy Greenwald \and Denizalp Goktas}
\affiliations{Simulacrum, New York City, NY, USA \quad \affil{*}Equal contribution.}
\reporttype{Technical Report}

\begin{document}
\maketitle

\begin{abstract}
  Time-series forecasting is central to many scientific and industrial domains, such as energy systems, climate modeling, finance, and retail. 
  While forecasting methods have evolved from classical statistical models to 
  automated, and neural approaches, the surrounding software ecosystem remains anchored to the traditional Python numerical stack. 
  Existing libraries rely on interpreter-driven execution and object-oriented abstractions, limiting composability, large-scale parallelism, and integration with modern differentiable and accelerator-oriented workflows.
  Meanwhile, today's forecasting increasingly involves large collections of heterogeneous time series data, irregular covariates, and frequent retraining, placing new demands on scalability and execution efficiency. 
  JAX offers an alternative paradigm to traditional stateful numerical computation frameworks based on pure functions and program transformations such as just-in-time compilation and automatic vectorization, enabling end-to-end optimization across CPUs, GPUs, and TPUs. 
  However, this modern paradigm has not yet been fully incorporated into the design of forecasting systems.
  We introduce \emph{Chronax}, a JAX-native time-series forecasting library that rethinks forecasting abstractions around functional purity, composable transformations, and accelerator-ready execution. 
  By representing preprocessing, modeling, and multi-horizon prediction as pure JAX functions, Chronax enables scalable multi-series forecasting, model-agnostic conformal uncertainty quantification, and seamless integration with modern machine learning and scientific computing pipelines. We provide access to our code on GitHub: \repo.
\end{abstract}

\section{Introduction}

Time-series forecasting plays a crucial role in a wide range of scientific research and industrial and operational domains, including energy management (e.g. \cite{uremovic2022new, chou2018forecasting}), climate modeling (e.g. \cite{ray2021time, adedotun2020modelling}), finance (e.g. \cite{dingli2017financial, nasir2025hybrid}), infrastructure (e.g. \cite{adekunle2021predictive, el2025smart}) and retail (e.g. \cite{oliveira2024evaluating, haque2023retail}). 
Over the past decade, the field has progressed through multiple methodological transitions. Early work was dominated by classical statistical forecasting models (e.g., ARIMA \cite{niknam2025comparing}), which rely on strong stationarity and linearity assumptions \cite{makridakis2023statistical, montero2021principles}.
As forecasting problems grew in scale and complexity, these approaches expanded into machine-learning-based methods, including tree-based models, regression-based approaches, and ensemble techniques that leverage feature engineering and cross-series information to model nonlinear relationships and incorporate exogenous variables \cite{lim2021temporal, zhou2021informer}.
More recently, the field has further evolved toward deep-learning-based forecasting methods, such as recurrent neural networks, temporal convolutional networks, and transformer-based architectures, which aim to learn shared temporal representations directly from data and scale to large collections of heterogeneous time series \cite{makridakis2023statistical}.
Despite this evolution, many practitioners still resort to statistical forecasting methods as a first resort when approaching forecasting tasks, and much of the tooling that practitioners rely on remains anchored to a traditional numerical stack. 

Libraries such as StatsForecast \cite{garza2022statsforecast} demonstrate that NumPy, Numba, and com\-pi\-la\-tion-based extensions can accelerate classical forecasting models by reducing interpreter overhead and enabling fast batch training and inference across large numbers of time series. 
However, the nature and scale of forecasting applications is changing. 
This change is driven by the increasing availability of high-frequency and large-scale time-series, as well as the growing integration of forecasting models into operational, planning, and automated decision-making systems that require frequent retraining and low-latency execution \cite{montero2021principles, makridakis2023statistical}.
Many of today's forecasting systems require modeling collections of time series---often thousands at once---with heterogeneous behaviors, irregular covariates, and frequent retraining cycles.
Existing forecasting libraries inherit several structural constraints of a past era, including their fundamentally interpreter-driven execution models, limited support for parallel, fully compositional, and differentiable workflows, and design assumptions centered on single-series or small-batch forecasting, which limit their scalability in modern large-scale forecasting settings \cite{makridakis2023statistical, montero2021principles}.

Existing statistical forecasting libraries suffer from three major shortcomings. First, they struggle to efficiently forecast large collections of time series simultaneously, an increasingly common requirement in scientific and industrial applications. This limitation arises because many libraries provide limited support for parallelized computing, making it difficult to exploit parallel execution across thousands of independent series \cite{meisenbacher2022review}.
Second, \sdeni{}{most} forecasting pipelines offer limited support for scalable, hardware-efficient execution. Most implementations remain structured around Python interpreter-driven control flow and CPU-centric numerical routines; this reliance on interpreter-driven semantics restricts the application of compiler-guided optimizations and accelerator-backed execution \cite{siebert2021systematic}.
Third, integrating statistical forecasting libraries into modern differentiable and compositional machine learning workflows remains challenging. Existing tools frequently rely on object-oriented interfaces, which complicate functional composition and hinder seamless integration with differentiable or accelerator-oriented systems \cite{olivares2022hierarchicalforecast}.

Fundamentally, these problems stem from the limitations of the numerical computing libraries on top of which these statistical libraries are built.
In contrast to other numerical computing libraries, JAX offers a fundamentally different lens for constructing numerical software in Python---a transformation-oriented framework. 
JAX expresses computation as pure functions and transforms them using mechanisms such as \texttt{jit} (compilation), \texttt{vmap} (automatic vectorization), \texttt{lax.scan} (loop fusion), and \texttt{jax.grad} (automatic differentiation), thereby facilitating effortless parallelization and autodifferentiation through the entire training pipeline.
enabling the same Python code to target CPU, GPU, or TPU backends.
This model has reshaped workflows in reinforcement learning, optimization, and scientific simulation by allowing entire pipelines, not just isolated kernels, to be optimized through program transformations. 
Time-series forecasting, however, has not yet fully incorporated these paradigms into its foundational design.

To address the aforementioned issues with existing we introduce Chronax, a JAX-native forecasting library designed to explore what a forecasting system looks like when built from first principles around functional purity, composable transformations, and accelerator-ready execution. 
Rather than porting existing CPU-oriented forecasting code, Chronax rethinks the structure and interface of forecasting itself. 
In Chronax, preprocessing operations, model computations, and multi-step prediction routines are represented as pure functions, enabling them to be transformed, compiled, vectorized, or composed without altering their semantics. 
Thus, Chronax provides a clean computational foundation for scalable forecasting workflows that align with modern machine learning and numerical computing practices.

\subsection{Related Work}

Recent forecasting libraries reflect a growing emphasis on scalability and modularity in time series forecasting.
HierarchicalForecast \cite{olivares2022hierarchicalforecast} is designed to generate coherent forecasts across hierarchies, 
meaning methods that enforce reconciliation constraints so that forecasts produced at lower levels aggregate to higher-level forecasts.
By correcting independently generated predictions to satisfy summation or grouping relationships, these approaches ensure coherence across all levels of the hierarchy. 
ForeTiS \cite{eiglsperger2023foretis} proposes an extensible experimental and benchmarking environment that supports both classical and machine learning forecasting models, enabling systematic exploration of design choices. 
Tsururu \cite{kostromina2025tsururu} introduces a modular architecture that combines preprocessing operations, forecasting strategies, and model families, and supports global and multivariate approaches for industrial applications. 
Although these systems broaden the modeling landscape, they remain built on the standard Python/NumPy paradigm, meaning their abstractions, execution models, and performance characteristics are ultimately shaped by the Python interpreter rather than by compiler-guided numerical transformations.

\subsection{Contributions}
This work introduce Chronax, a JAX-native times-series forecasting library, and thereby makes the following key contributions:

\begin{enumerate}
    \item \textbf{A functional
    forecasting framework:} 
    Chronax is built around pure functions to be compatible with JAX's transformation system.
    This stateless design enables effortless parallelization across devices, unlocking JAX's computational speed, while also ensuring transparent computation, clearer data flow, and seamless integration with differentiable workflows and scientific machine-learning pipelines.
    
    \item \textbf{A unified abstraction for scalable multi-series forecasting:}
    Chronax structures forecasting routines to be automatically vectorizable across large collections of time series using \texttt{vmap}.
    This eliminates the need for multiprocessing, Python loops, or hand-engineered batching logic, offering a general abstraction suitable for modern large-scale forecasting tasks.
    
    \item \textbf{An end-to-end differentiable pipeline:} 
    All components---including preprocessing, model evaluation, feature transformations, and multi-horizon prediction---are implemented as JAX-compatible functions. 
    This not only ensures that JAX transformations 
    can be applied consistently throughout the forecasting workflow, it further enables end-to-end autodifferentiation, allowing gradients to be propagated through the entire pipeline for optimization, learning, and hypergradient-based optimization extensions.
    
    \item \textbf{Model-agnostic conformal prediction integrated into the forecasting process:} 
    Chronax incorporates conformal inference methods in a manner consistent with JAX’s functional semantics, enabling uncertainty quantification to benefit directly from the same transformation and compilation tools.
    
    \item \textbf{A modern reinterpretation of classical forecasting abstractions:} By departing from mixed Python–Numba–C++ implementations and embracing a unified JAX-based architecture, Chronax provides a cleaner and more maintainable foundation for extensibility and experimentation. %
\end{enumerate}

\section{Background}\label{sec:background}
\subsection{Time-series Forecasting}

A \mydef{(time-series) forecasting task} $\forecastprob \doteq (\contextlen, \forecastlen, \numcovars, \numtargets, (\covarset[\covar])_{\covar \in [\numcovars]}, (\targetset[\target])_{\target \in [\numtargets]}, \covarts, \targetts)$ comprises a \mydef{context length} $\contextlen \in \N$, a \mydef{forecast horizon} $\forecastlen \in \N$, 
$\numtargets \in \N$ \mydef{target time-series} $\targetts = (\targetts[1], \hdots, \targetts[\numtargets])^T$,
where for each variate $\target \in \targets$, entries of $\targetts[\target] \in \targetset[\target]^{\contextlen}$ take values $\targetts[\target][t]$, for $t \in [\contextlen]$, from a \mydef{set of target values} %
$\targetset[\target] \subseteq \R$,
and $\numcovars \in \N$ \mydef{covariate time-series} $\covarts = (\covarts[1], \hdots, \covarts[\numcovars])^T$,
where for each covariate $\covar \in \covars$, $\covarts[\covar] \in \covarset[\covar]^{\contextlen + \forecastlen}$ takes values $\covarts[\covar][t]$, for $t \in [\contextlen + \forecastlen]$, from a \mydef{set of covariate values} %
$\covarset[\covar] \subseteq \R$.
We denote the \mydef{joint set of covariate values} by $\covarset^{\contextlen + \forecastlen} \doteq \bigtimes_{\covar \in \covars} \covarset[\covar]^{\contextlen + \forecastlen}$ and the \mydef{joint set of target variate values} by $\targetset^{\contextlen} \doteq \bigtimes_{\target \in \targets} \targetset[\target]^{\contextlen}$.
A forecasting task $\forecastprob$ is said to be \mydef{univariate} (resp.\ \mydef{multivariate}) iff $\numtargets = 1$ ($\numtargets > 1$). A forecasting task $\forecastprob$ is said to be \mydef{unconditional} (resp.\ \mydef{conditional}) iff $\numcovars = 0$ (resp.\ $\numcovars > 0$). 

Given a forecasting task $\forecastprob$, a \mydef{(point) forecast} is a matrix $\forecastts \doteq (\forecastts[1], \hdots, \forecastts[\numtargets])^T$ s.t.\ for all target variates $\target \in \targets$, $\forecastts[\target] \in \targetset[\target]^{\forecastlen}$
represents forecasted values of variate $\target$ for $\forecastlen$ time steps. 
A \mydef{(point) forecasting model} 
is a mapping $\forecaster: \covarset^{\contextlen + \forecastlen} \times \targetset^{\contextlen} \to \targetset^{\forecastlen}$ whose output is a forecast for $\forecastprob$: i.e., $\forecaster(\covarts, \targetts) \doteq (\forecaster[1](\covarts, \targetts), \hdots, \forecaster[\numtargets](\covarts, \targetts))^T = \forecastts$.
More generally, a \mydef{probabilistic forecasting model} is a mapping $\stochforecaster: \covarset^{\contextlen + \forecastlen} \times \targetset^{\contextlen} \to \simplex(\targetset^{\forecastlen})$ s.t. $\stochforecaster(\covarts, \targetts)[\forecastts]$ denotes the realization probability of $\forecastts \in \targetset^{\forecastlen}$.

\subsection{Conformal Prediction for Time Series}
\label{sec:conformal}

Conformal prediction provides a distribution-free framework for constructing valid prediction intervals from empirical prediction errors computed on a held-out validation set \cite{shafer2008tutorial}.
Chronax provides two interval construction methods that differ in how they use the conformity scores: the \emph{conformal distribution} method, which computes conformity scores using absolute residuals to produce symmetric intervals; and the \emph{conformal signed} method, which retains error direction to produce asymmetric intervals.

\paragraph{Conformity Scores via Conformal Validation.}

We describe the construction of conformity scores for a single target variate $\target \in \targets$;
this same construction applies independently to each variate.
Given $\contextlen$ observations $\targetts[\target][1], \ldots, \targetts[\target][\contextlen]$ and a forecast horizon $\forecastlen$, we compute conformity scores using \mydef{conformal validation}, which can be understood as ``walk-forward" cross validation:

\begin{enumerate}
    \item \textbf{Partition the series into $\numwindows$ calibration windows.} 
We divide the time series into $\numwindows$ train-calibration splits, called \mydef{calibration windows}, each of which respects the temporal ordering.

The calibration windows are formed by partitioning the data, working backwards from the end.
The final $\forecastlen \numwindows$ observations are partitioned into $\numwindows$ consecutive, non-overlapping calibration blocks of size $\forecastlen$.
Each window's \mydef{training set} then contains all observations preceding its calibration block, so
$\targetts[\target][1], \ldots, \targetts[\target][{,\timestep_{\window}}]$, with $\window \in \windows$ and $\timestep_{\window} = \contextlen - (\numwindows + 1 - \window) \cdot \forecastlen$, and the \mydef{calibration set} contains the subsequent $\forecastlen$ observations $\targetts[\target][{,\timestep_{\window}+1}], \ldots, \targetts[\target][{,\timestep_{\window}+\forecastlen}]$. 
Each successive training set grows by exactly $\forecastlen$ observations, and all observations preceding the first calibration block are included in every window's training set.

\newcommand{\mytimes}{\mathbin{\scriptstyle\times}}

\begin{figure}[ht]
\centering
\begin{tikzpicture}
    \matrix (m) [
        matrix of math nodes,
        left delimiter=[,
        right delimiter={]},
        column sep=0.5em,
        row sep=0.5em,
        nodes={anchor=center, minimum width=1.5em}
    ]
    {
        \circ  & \circ  & \mytimes      & \mytimes      & \mytimes      & \mytimes     \\
        \circ  & \circ  & \circ  & \circ  & \circ  & \circ  & \mytimes      & \mytimes      & \mytimes      & \mytimes      \\
        \circ  & \circ  & \circ  & \circ  & \circ  & \circ  & \circ  & \circ  & \circ  & \circ  & \mytimes      & \mytimes      & \mytimes      & \mytimes      \\
    };
\end{tikzpicture}

\caption{Conformal validation assuming $T = 14$, $\numwindows = 3$, and $\forecastlen = 4$. For $\window = 1$, $\timestep_{\window} = 2$; for $\window = 2$, $\timestep_{\window} = 6$; for $\window = 3$, $\timestep_{\window} = 10$. Circles ($\circ$) denote training observations; Xs ($\mytimes$) denote the $\forecastlen$ calibration steps used for conformal scoring. Each successive calibration window's training set grows by $\forecastlen$ observations.}
\end{figure}

    \item \textbf{Generate forecasts for each window.}
    For each calibration window $\window$, we fit the forecasting model $\forecaster[\target]$ on the training data and produce forecasts for the next $\forecastlen$ time steps $\forecastts[\target][{,\timestep_{\window}+1}], \ldots, \forecastts[\target][{,\timestep_{\window}+\forecastlen}]$. As these time steps are defined relative to a calibration window $\window$, we refer to them as ``calibration steps."
    We repeat this process across multiple windows, with the intent of capturing the model's error distribution across time.

    \item \textbf{Compute conformity scores.}
    We record the signed residuals as conformity scores:
    \begin{equation}
    \label{eq:confscore}
        \confscore[\window][\calibstep] \doteq \targetts[\target][{,\timestep_{\window}+\calibstep}] - \forecastts[\target][{,\timestep_{\window}+\calibstep}], \quad \calibstep = 1, \ldots, \forecastlen .
    \end{equation}
\end{enumerate}

This procedure yields a $\numwindows \times \forecastlen$ conformity score matrix $\confscore$, where rows index calibration windows and columns index forecast steps $\calibstep = 1, \ldots, \forecastlen$.

\paragraph{(Symmetric) Conformal Prediction Method.}

We use conformity scores to compute prediction intervals.

Given a point forecast $\forecastts[\target][{,\contextlen+\calibstep}]$ produced on the full observed context $\targetts[\target][,1], \ldots, \targetts[\target][{\contextlen}]$, we construct a prediction interval at future step $\calibstep = 1, \ldots, \forecastlen$ from the absolute conformity scores $|\confscore[1][\calibstep]|, \ldots, |\confscore[\numwindows][\calibstep]|$.

We then form an empirical distribution $\plausibleset[\calibstep]$ from the set of plausible realizations at $\calibstep$ as follows:
\begin{equation}
\label{eq:plausibleset}
    \plausibleset[\calibstep] \doteq \left\{ \forecastts[\target][{,\contextlen+\calibstep}] \pm |\confscore[\window][\calibstep]| \;:\; \window \in \windows \right\}.
\end{equation}
This set contains $2\numwindows$ perturbed values of the point forecast $\forecastts[\target][{,\contextlen+\calibstep}]$, obtained by perturbing $\forecastts[\target][{,\contextlen+\calibstep}]$ by observed error magnitudes at calibration step $\calibstep$.
Finally, a $(1-\coveragelvl)$ \mydef{prediction interval} $\confinterval$ is obtained by taking quantiles of the empirical distribution $\plausibleset[\calibstep]$, as follows:
\begin{equation}
\label{eq:predinterval}
    \confinterval = \left[\quantilefn[\coveragelvl/2]\!\left(\plausibleset[\calibstep]\right),\; \quantilefn[1-\coveragelvl/2]\!\left(\plausibleset[\calibstep]\right)\right],
\end{equation}
where $\quantilefn[p](\cdot)$ denotes the $p$-th quantile, for $p \in [0, 1]$.
Because $\plausibleset[\calibstep]$ is symmetric around $\forecastts[\target][{,\contextlen+\calibstep}]$ by construction, the resulting interval is likewise symmetric.

\paragraph{Exchangeability of Conformity Scores.}

A sequence of random variables $\targetYvariable[1], \ldots, \targetYvariable[T]$ is \mydef{exchangeable} if the joint distribution of the random variables is invariant under permutations. 
Assuming exchangeability, the ranks of the entries in a set of conformal scores are uniformly distributed.
As a result, given a tail probability (or error rate) $\alpha \in (0, 1)$, conformal prediction intervals achieve $(1 - \alpha)$-coverage, meaning
\begin{equation}
    \Pr \left( \targetYvariable[\calibstep] \in \confinterval \right) \ge 1 - \alpha .
\end{equation}

The smallest nonzero tail probability that can be represented by a set of $N$ exchangeable conformal scores is of order $\nicefrac{1}{N}$,
because each discrete score contributes mass $\nicefrac{1}{N}$ to the empirical distribution.
Therefore, conformal prediction can only provide finite-sample guarantees for coverage levels $1 - \alpha$, when $\alpha \gtrsim \nicefrac{1}{N}$.

Typical conformal scores produced from time-series data violate the exchangeability assumption for various reasons, such as temporal dependence (the future depends on the past) and non-stationarity.
Nonetheless, \citet{barber2023conformal} show that coverage loss can be bounded by the total variation distance between the true (non-exchangeable) score distribution and the nearest exchangeable one.

\paragraph{Signed Conformal Prediction Method.}

The symmetric conformal prediction method is based on the absolute values of the residuals, and thus discards error direction.
When a forecaster exhibits systematic bias, it can be useful to retain signed residuals, and produce intervals that reflect this asymmetry.
Define the set of $\numwindows$ signed plausible realizations at horizon $\calibstep$ as follows:
\begin{equation}
\label{eq:plausiblesetsigned}
    \plausibleset[\calibstep][\mathrm{signed}] \doteq \left\{\forecastts[\target][{,\contextlen+\calibstep}] + \confscore[\window][\calibstep] \;:\; \window \in \windows\right\}.
\end{equation}
Given this ``signed" set of values, the conformal signed method uses prediction intervals constructed analogously to~\eqref{eq:predinterval}:
\begin{equation}
\label{eq:predintervalsigned}
    \left[\quantilefn[\coveragelvl/2]\!\left(\plausibleset[\calibstep][\mathrm{signed}]\right),\; \quantilefn[1-\coveragelvl/2]\!\left(\plausibleset[\calibstep][\mathrm{signed}]\right)\right].
\end{equation}
When the signed scores are not centered at zero, the $\coveragelvl/2$ and $1-\coveragelvl/2$ quantiles of $\plausibleset[\calibstep][\mathrm{signed}]$ are not equidistant from the point forecast $\forecastts[\target][{,\contextlen+\calibstep}]$, yielding asymmetric intervals.

Each of the $\numwindows$ signed scores in Eq.~\eqref{eq:predintervalsigned} contributes mass $\nicefrac{1}{\numwindows}$ to the empirical distribution, so the $\nicefrac{\coveragelvl}{2}$-quantile cannot resolve probabilities below $\nicefrac{1}{\numwindows}$: i.e., it is necessary that $\nicefrac{\coveragelvl}{2} \ge \nicefrac{1}{\numwindows}$, giving $\coveragelvl \ge \nicefrac{2}{\numwindows}$.
As the test point itself occupies one of the exchangeable rank slots, the inequality is more accurately expressed as $\coveragelvl \ge \nicefrac{2}{\numwindows + 1}$, yielding a minimum coverage level of approximately $1 - \nicefrac{2}{\numwindows}$.
In contrast, the symmetric method effectively uses $2\numwindows$ points of mass $\nicefrac{1}{2\numwindows}$ each, so $\nicefrac{\coveragelvl}{2} \ge \nicefrac{1}{2\numwindows}$, yielding $\coveragelvl \ge \nicefrac{1}{\numwindows}$; the same rank-slot correction yields a minimum coverage level of approximately $1 - \nicefrac{1}{\numwindows}$.
Therefore, the finite-sample coverage guarantee degrades faster for the signed method.

\paragraph{Efficient Implementation via \texttt{vmap}.}

A naive implementation of walk-forward cross validation requires $\numwindows$ sequential model fits.
Chronax parallelizes this computation using \texttt{jax.vmap} to vectorize all $\numwindows$ windows simultaneously, yielding significant speedups on GPU/TPU hardware.

\section{Design Principles}

\definecolor{codegreen}{rgb}{0,0.6,0}
\definecolor{codegray}{rgb}{0.5,0.5,0.5}
\definecolor{codepurple}{rgb}{0.58,0,0.82}
\definecolor{backcolour}{rgb}{0.97,0.97,0.97}

\lstset{
    language=Python,
    backgroundcolor=\color{backcolour},
    basicstyle=\ttfamily\footnotesize,
    keywordstyle=\color{blue},
    commentstyle=\color{codegreen},
    stringstyle=\color{codepurple},
    showstringspaces=false,
    breaklines=true,
    columns=fullflexible,
    keepspaces=true,
    frame=single,
    xleftmargin=0pt,
    xrightmargin=0pt,
    aboveskip=0.5em,
    belowskip=0.5em,
    float=htbp
}

\subsection{Library Architecture}

The overarching goal in building Chronax was to create a JAX-native forecasting library whose numerical routines, array operations, and compilation pathways work naturally with JAX. Chronax exposes a class-based forecasting interface built around shared estimator methods such as \texttt{fit()}, \texttt{predict()}, \texttt{forecast()}, and \texttt{forward()}, while delegating much of the numerical computation to JAX-compatible kernels and tensor operations.

Chronax employs a lightweight mixed package structure that prioritizes direct access to public models while grouping larger model families into dedicated subpackages. Smaller forecasters are implemented as standalone modules, whereas larger families such as ARIMA, ETS, TBATS, Theta, and MFLES reside in dedicated subdirectories. This organization keeps the public API compact while localizing model-specific internals.

\subsubsection{Directory Organization}

\begin{itemize}
    \item \textbf{Model modules and family packages}: Public forecasting classes are exposed from \path{chronax.models}. 
    The implementation uses both standalone modules (for example, \path{holt.py}, \path{holt_winters.py}, \path{naive.py}, \path{garch.py}) and family subpackages (for example, \path{chronax/models/arima/}, \path{chronax/models/ets/}, \path{chronax/models/tbats/}, \path{chronax/models/theta/}, and \path{chronax/models/mfles/}).
    Typical imports follow the packaged interface: e.g., \\
    \texttt{from chronax.models import HoltWinters}.

    \item \textbf{Core infrastructure}:
    \begin{itemize}
        \item \texttt{chronax/models/base\_forecaster.py}: abstract base class defining the shared estimator interface and conformal-interval utilities
        \item \texttt{chronax/utils/conformal\_intervals.py}: 
        configuration container for conformal interval parameters
        \item \texttt{chronax/utils/loss\_functions.py}: deterministic and probabilistic forecast metrics implemented with JAX arrays
        \item \texttt{chronax/utils/utils.py}: shared numerical helpers used across forecasting models
    \end{itemize}
\end{itemize}

\subsubsection{Architectural Principles}

\begin{enumerate}
    \item \textbf{Unified estimator interface}: 
    Public forecasting models inherit from \texttt{BaseForecaster} and implement a common interface centered on \texttt{fit()}, \texttt{predict()}, and \texttt{forecast()}. The base class also provides default behavior for \texttt{forward()}, conformity score computation, and conformal interval post-processing.

    \item \textbf{JAX-oriented numerical implementation}: 
    Chronax relies heavily on JAX arrays and transformations such as \texttt{jax.numpy}, \texttt{jax.jit}, \texttt{jax.vmap}, and \texttt{jax.lax} to express model fitting, forecasting, and batch evaluation in hardware-accelerated form. 
    The degree of compilation and vectorization varies by model, but JAX arrays and transformations (e.g., \texttt{jax.numpy}, \texttt{jax.jit}, \texttt{jax.lax}, \texttt{jax.vmap}) are used by every forecasting model in the library.

    \item \textbf{Minimal inheritance hierarchy}: 
    Most public forecasters inherit directly from \texttt{BaseForecaster} rather than from deeper family-specific abstract layers. Family-level complexity is typically handled within model modules or subpackages instead of through a large inheritance tree.

    \item \textbf{Compilation-aware API design}: 
    Chronax is designed to work well with JAX compilation, using a mix of the \texttt{@jax.jit} decorator and \texttt{static\_argnums} and \texttt{static\_argnames}, depending on the model.
\end{enumerate}

\subsection{Models Implemented}

Chronax currently exposes 28 public forecasting model classes through \path{chronax.models}, spanning eight model families. 
This count excludes non-forecasting helper exports such as \texttt{BatchedForecaster} as well as private helper functions (for example, names prefixed by an underscore).
Table~\ref{tab:chronax-models} summarizes the current model collection and its public forecasting capabilities.

\begin{longtable}{p{3.0cm} c c c c c}
\caption{Chronax model capabilities by family.}
\label{tab:chronax-models} \\
\toprule
Model & \shortstack{Point\\Forecast} & \shortstack{Probabilistic\\Forecast} & \shortstack{In-sample\\Fitted Values} & \shortstack{Probabilistic\\Fitted Values} & \shortstack{Exogenous\\Features} \\
\midrule
\endfirsthead
\toprule
Model & \shortstack{Point\\Forecast} & \shortstack{Probabilistic\\Forecast} & \shortstack{In-sample\\Fitted Values} & \shortstack{Probabilistic\\Fitted Values} & \shortstack{Exogenous\\Features} \\
\midrule
\endhead
\bottomrule
\endfoot
\multicolumn{6}{l}{\textbf{Automatic Forecasting}} \\
\midrule
AutoARIMA   & $\checkmark$ & $\checkmark$ & $\checkmark$ & $\checkmark$ & $\checkmark$ \\
AutoETS     & $\checkmark$ & $\checkmark$ & $\checkmark$ & $\checkmark$ &  \\
AutoTheta   & $\checkmark$ & $\checkmark$ & $\checkmark$ & $\checkmark$ &  \\
AutoMFLES   & $\checkmark$ & $\checkmark$ & $\checkmark$ &  & $\checkmark$ \\
AutoTBATS   & $\checkmark$ & $\checkmark$ & $\checkmark$ & $\checkmark$ &  \\
AutoCES     & $\checkmark$ & $\checkmark$ & $\checkmark$ &  &  \\
\addlinespace
\multicolumn{6}{l}{\textbf{ARIMA Family}} \\
\midrule
ARIMA       & $\checkmark$ & $\checkmark$ & $\checkmark$ & $\checkmark$ & $\checkmark$ \\
\addlinespace
\multicolumn{6}{l}{\textbf{Theta Family}} \\
\midrule
Theta       & $\checkmark$ & $\checkmark$ & $\checkmark$ & $\checkmark$ &  \\
\addlinespace
\multicolumn{6}{l}{\textbf{Multiple Seasonalities \& Decomposition}} \\
\midrule
MFLES       & $\checkmark$ & $\checkmark$ & $\checkmark$ &  & $\checkmark$ \\
TBATS       & $\checkmark$ & $\checkmark$ & $\checkmark$ & $\checkmark$ &  \\
MSTL        & $\checkmark$ & $\checkmark$ & $\checkmark$ & $\checkmark$ &  \\
STL         & $\checkmark$ & $\checkmark$ & $\checkmark$ & $\checkmark$ &  \\
\addlinespace
\multicolumn{6}{l}{\textbf{Volatility Models}} \\
\midrule
GARCH       & $\checkmark$ & $\checkmark$ & $\checkmark$ & $\checkmark$ &  \\
\addlinespace
\multicolumn{6}{l}{\textbf{Baseline Models}} \\
\midrule
Hist.\ Avg. & $\checkmark$ & $\checkmark$ & $\checkmark$ & $\checkmark$ &  \\
Naive       & $\checkmark$ & $\checkmark$ & $\checkmark$ & $\checkmark$ &  \\
Seasonal Naive & $\checkmark$ & $\checkmark$ & $\checkmark$ & $\checkmark$ &  \\
Window Avg. & $\checkmark$ & $\checkmark$ &  &  &  \\
Seasonal Window Avg. & $\checkmark$ & $\checkmark$ &  &  &  \\
RWD         & $\checkmark$ & $\checkmark$ & $\checkmark$ & $\checkmark$ &  \\
\addlinespace
\multicolumn{6}{l}{\textbf{Exponential Smoothing}} \\
\midrule
ETS         & $\checkmark$ & $\checkmark$ & $\checkmark$ & $\checkmark$ &  \\
Holt        & $\checkmark$ & $\checkmark$ & $\checkmark$ & $\checkmark$ &  \\
Holt--Winters & $\checkmark$ & $\checkmark$ & $\checkmark$ & $\checkmark$ &  \\
SES         & $\checkmark$ & $\checkmark$ & $\checkmark$ &  &  \\
Seasonal ES & $\checkmark$ & $\checkmark$ & $\checkmark$ &  &  \\
\addlinespace
\multicolumn{6}{l}{\textbf{Sparse / Intermittent Demand}} \\
\midrule
ADIDA       & $\checkmark$ & $\checkmark$ & $\checkmark$ & $\checkmark$ &  \\
CrostonClassic & $\checkmark$ & $\checkmark$ & $\checkmark$ & $\checkmark$ &  \\
IMAPA       & $\checkmark$ & $\checkmark$ & $\checkmark$ & $\checkmark$ &  \\
TSB         & $\checkmark$ & $\checkmark$ & $\checkmark$ & $\checkmark$ &  \\
\end{longtable}

{\footnotesize
\textit{Notes.}
A checkmark indicates that the corresponding capability is exposed by the current public Chronax model API.
``Probabilistic forecast'' and ``probabilistic fitted values'' may arise from native analytical intervals, Monte Carlo procedures, Gaussian approximations, conformal intervals, or combinations thereof, depending on the model.
Here, ``probabilistic fitted values'' means that the model exposes uncertainty bounds for an in-sample fitted or reconstructed series through its public API.
``Exogenous features'' indicates that the public interface accepts aligned regressors at fit time and, where applicable, future covariates at prediction time.

The forecaster classes summarized above inherit from \texttt{BaseForecaster}, which defines the core forecasting interface:

\begin{lstlisting}[xrightmargin=-3.33mm, caption={BaseForecaster interface}]
class BaseForecaster:
    # Core forecasting methods
    def fit(self, y, X=None) -> self
    def predict(self, h, X_future=None, level=None) -> dict
    def forecast(self, y, h, X=None, X_future=None, level=None, fitted=False) -> dict
    def forward(self, y, h, X=None, X_future=None, level=None, fitted=False) -> dict
\end{lstlisting}

\begin{lstlisting}[caption={Typical Chronax workflow}]
from chronax.models import HoltWinters
from chronax.utils.conformal_intervals import ConformalIntervals

# Initialize model with conformal intervals
model = HoltWinters(
    season_length=12,
    error_type='M',
    season_type='A',
    conformal_params=ConformalIntervals(n_windows=5, h=12)
)

# Fit stores the Holt-Winters learned parameters for y_train on the model object. If conformal_params is specified, it stores the conformity scores in model.
model.fit(y_train) 

# Predict runs a stored model to produce h-step forecasts as well as confidence intervals if level is specified.
result = model.predict(h=12, level=[80, 95])

# Forecast is an alias for first fitting a model and then returning a forecast (with confidence intervals if level is specified). (This fit is not saved in the model object, so this method is stateless.)
result = model.forecast(y_train, h=12, level=[80, 95])

# Forward runs a model that was previously fit on y_train on y_new (with confidence intervals if level is specified).
result = model.forward(y_new, h=12, level=[80, 95])
\end{lstlisting}

This interface follows the scikit-learn-style \texttt{fit}/\texttt{predict} estimator API, while accommodating forecasting-specific methods and interval utilities.  %
The key methods are:

\begin{itemize}
    \item \texttt{fit(y, X)}: fits the model to training data, stores the learned model parameters in \texttt{model\_}, and returns \texttt{self}.
    For convenience, the learned parameters are saved in the model class, but if the user prefers a fully stateless execution, \texttt{fit()} also returns a copy of model with the learned parameters.

    \item \texttt{predict(h, X, level)}: generates $h$-step forecasts from an already fitted model, optionally returning interval bounds when supported by the model and requested through \texttt{level}.

    \item \texttt{forecast(y, h, ...)}: performs fit-plus-predict in one call (without retaining model state).

    \item \texttt{forward(y, h, \ldots)}: by default, this method is the same as \texttt{forecast(y, h, \ldots)}. But some models override the default to forecast from a new history \texttt{y} reusing an earlier fitted model.
    Details differ by model.    
\end{itemize}

\section{Experiments}

To rigorously assess the performance and scalability of the Chronax forecasting framework relative to the StatsForecast baseline, we designed a comprehensive evaluation procedure. 
This procedure is designed to ensure a fair, "neutral ground" comparison by controlling for data generation, hardware states, and execution overheads across both the JAX-based (Chronax) and Numba-compiled (StatsForecast) environments.

Chronax' evaluation framework uses a multi-environment architecture, meaning multiple isolated setups, to avoid conflicts between JAX and Numba dependencies. 
A central orchestrator spawns each model as an isolated worker process, ensuring a clean execution environment.

Our experiments comprise head-to-head comparisons between candidate models in Chronax (e.g., \texttt{WindowAverage}, \texttt{ETS}, \texttt{TSB}), and their algorithmic equivalents in StatsForecast.
All experiments were conducted using a unified \texttt{fit\_predict} interface wrapped around both libraries.
Data transfer times (e.g., moving NumPy arrays to JAX DeviceArrays) are handled prior to the timing loop to isolate inference speed.

\subsection{Metrics}
\label{sec:metrics}

We collect six primary metrics to evaluate both computational efficiency and predictive performance:

\begin{enumerate}
    \item \textbf{Cold Start Time ($T_{\text{cold}}$)}:
    Measures the total time for the first execution, capturing the one-time costs of JIT compilation (XLA for JAX, LLVM for Numba) and initialization.
    $$ T_{\text{cold}} = t_{\text{end}, 0} - t_{\text{start}, 0} $$

    \item \textbf{Warm Start Time ($T_{\text{warm}}$)}:
    Represents the steady-state throughput of the model. After the initial cold start, the model is executed for $N=5$ iterations. The average duration is recorded. Crucially, for JAX models, we enforce explicit synchronization via \texttt{.block\_until\_ready()} to measure actual GPU computation time rather than asynchronous dispatch time.
    $$ T_{\text{warm}} = \frac{1}{N} \sum_{i=1}^{N} (t_{\text{end}, i} - t_{\text{start}, i}) $$
    
    \item \textbf{MAPE}:
    To verify that performance optimizations do not degrade predictive quality, we calculate the Mean Absolute Percentage Error on a holdout test set of length $H=24$.
    $$ \text{MAPE} = \frac{100\%}{H} \sum_{t=1}^{H} \left| \frac{y_{\text{true}}^{(t)} - y_{\text{pred}}^{(t)}}{y_{\text{true}}^{(t)}} \right| $$

   \item \textbf{MAE}:
    Evaluates the mean absolute error, providing a linear penalty for forecast deviations. It maintains the original scale of the data, offering a highly interpretable measure of the average error magnitude.
    $$ \text{MAE} = \frac{1}{H} \sum_{t=1}^{H} \left| y_{\text{true}}^{(t)} - y_{\text{pred}}^{(t)} \right| $$

    \item \textbf{RMSE}:
    Calculates the root mean squared error. By squaring the deviations before averaging, this metric disproportionately penalizes larger forecasting errors, making it particularly sensitive to outliers and high-variance periods in stochastic datasets.
    $$ \text{RMSE} = \sqrt{ \frac{1}{H} \sum_{t=1}^{H} \left( y_{\text{true}}^{(t)} - y_{\text{pred}}^{(t)} \right)^2 } $$

    \item \textbf{MASE}:
    Measures the mean absolute scaled error to provide a scale-independent evaluation across differing data-generating processes. To strictly align with our evaluation framework's execution, the model's MAE is scaled by the mean absolute error of a one-step-ahead naive forecast computed directly over the holdout test set.
    $$ \text{MASE} = \frac{ \frac{1}{H} \sum_{t=1}^{H} \left| y_{\text{true}}^{(t)} - y_{\text{pred}}^{(t)} \right| }{ \frac{1}{H-1} \sum_{t=2}^{H} \left| y_{\text{true}}^{(t)} - y_{\text{true}}^{(t-1)} \right| } $$
    
\end{enumerate}

\subsection{Datasets}

We evaluate the performance of our models on three real-world time series datasets with different ranges:

\begin{enumerate}
    \item \textbf{Airline Passengers Dataset}
    \cite{milleanno2024airpassengers}: This dataset is the smallest among our three selected datasets, recording only 144 observations. It includes monthly US airline passengers from 1949 to 1960 with two columns: month and number of passengers. The month column starts in January 1949 and ends in December 1960, formatted as yyyy-mm. The number of passengers is recorded as 3-digit integers, with minimum 104 and maximum 622.
    
    \item \textbf{Daily Female Births Dataset} \cite{dougcresswell1959dailybirths}: With 365 observations documented, this dataset is an example of mid-size time-series data. It records the total number of female births in California, USA during 1959 in two columns: date and number of births. The date column starts on January 1st, 1959, and ends on December 31st, 1959, with values increasing by one day in each row. The date column is formatted as yyyy-mm-dd. The number of births is recorded as 2-digit integers, with minimum 23 and maximum 73.

    \item \textbf{Room Temperatures Dataset} \cite{vitthalmadane2022ts}: This dataset includes 7056 observations. It lists room air temperatures recorded using an IOT device. In order to fit this dataset with our models, we removed the index column. After this modification, this dataset has two columns: datetime (date and hour at which each datum was recorded) and the \texttt{\#hourly\_temp} (the temperature of the air supplied to a room). The datetime column starts on January 4th, 2022 12 a.m. and ends on October 24th, 2022 11 p.m., with formatting as yyyy-mm-dd hh:mm:ss. This column increases by one hour with each row. The \texttt{\#hourly\_temp} column records real numbers up to three significant digits, with a minimum of 5.35 and maximum of 36.5.
\end{enumerate}

\begin{table}[h!]
\begin{center}
\caption{Datasets}
\begin{tabular}{|>{\raggedright\arraybackslash}p{3cm}||>{\raggedright\arraybackslash}p{3cm}|>{\raggedright\arraybackslash}p{3cm}|>{\raggedright\arraybackslash}p{4cm}|}
\hline
&\hyperlink{https://www.kaggle.com/datasets/brmil07/air-passengers-dataset}{Airline Passengers}&\hyperlink{https://www.kaggle.com/datasets/dougcresswell/daily-total-female-births-in-california-1959}{Daily Female Births}&\hyperlink{https://www.kaggle.com/datasets/vitthalmadane/ts-temp-1?select=MLTempDataset1.csv}{Room Temperatures}\\
\hline
\hline
Domain&Travel&Health&Physics\\
\hline
Variable Name&month&date&datetime\\
&number of passengers&number of births&number of \#hourly\_temp\\
\hline
Variable Units&yyyy-mm&yyyy-mm&yyyy-mm-dd hh:mm:ss\\
&3-digit integers&2-digit integers&real numbers\\
\hline
Number of Observations&144 observations&365 observations&7056 observations\\
\hline
\end{tabular}
\end{center}
\end{table}

\subsection{Results: Speed and Accuracy Analysis}

To evaluate the performance of Chronax relative to that of StatsForecast, we ran all models on these three datasets, collecting prediction qualities as determined by the aforementioned suite of metrics (MAPE, MAE, RMSE, and MASE; see Section~\ref{sec:metrics}).
In addition to this accuracy comparison, we also compare execution latency, measuring both cold start initialization and warm start execution times.

\subsubsection{Ratio Analysis}

\Cref{tab:means-medians} reports summary statistics collected from our experiments.\footnote{For comprehensive model-specific head-to-head comparisons and error metrics, see \Cref{sec:appendix}.}
These statistics are ratios, computed as the value achieved by Chronax divided by that of Statsforecast.
For latency metrics, 
values $<1.0\times$ indicate Chronax is faster. 
For accuracy metrics, values $<1.0\times$ indicate Chronax achieved lower error.

Because the mean is subject to distortion by extreme values, we report two statistics:
\begin{itemize}
    \item \textbf{Mean} \hspace{2mm} average value of all calculated model ratios
    \item \textbf{Median} \hspace{2mm} median value of all calculated model ratios
\end{itemize}

\begin{table}[hbt!]
\centering
\caption{Chronax / Statsforecast Ratios for all three datasets.}
\begin{tabular}{|l||cc|cc|cc|}
\hline
& \multicolumn{2}{c|}{\textbf{Airline Passengers}} & \multicolumn{2}{c|}{\textbf{Daily Female Births}} & \multicolumn{2}{c|}{\textbf{Room Temperature}} \\
\textbf{Metric} & \textbf{Mean} & \textbf{Median} & \textbf{Mean} & \textbf{Median} & \textbf{Mean} & \textbf{Median} \\
\hline
MAPE       & 1.11$\times$ & 1.00$\times$  & 1.01$\times$ & 1.00$\times$  & 0.95$\times$ & 1.00$\times$ \\
MAE        & 1.10$\times$ & 1.00$\times$  & 1.01$\times$ & 1.00$\times$  & 0.94$\times$ & 1.00$\times$ \\
RMSE       & 1.09$\times$ & 1.00$\times$  & 1.02$\times$ & 1.00$\times$  & 0.94$\times$ & 1.00$\times$ \\
MASE       & 1.10$\times$ & 1.00$\times$  & 1.01$\times$ & 1.00$\times$  & 0.94$\times$ & 1.00$\times$ \\
\hline
Cold Start Initialization & 44.54$\times$ & 10.19$\times$ & 19.32$\times$ & 9.23$\times$ & 5.82$\times$ & 1.77$\times$ \\
Warm Start Execution & 0.76$\times$ & 0.053$\times$ & 0.62$\times$ & 0.040$\times$ & 0.16$\times$ & 0.074$\times$ \\
\hline
\end{tabular}
\label{tab:means-medians}
\end{table}

\begin{figure*}[!htpb]
    \centering
    \includegraphics[width=\textwidth]{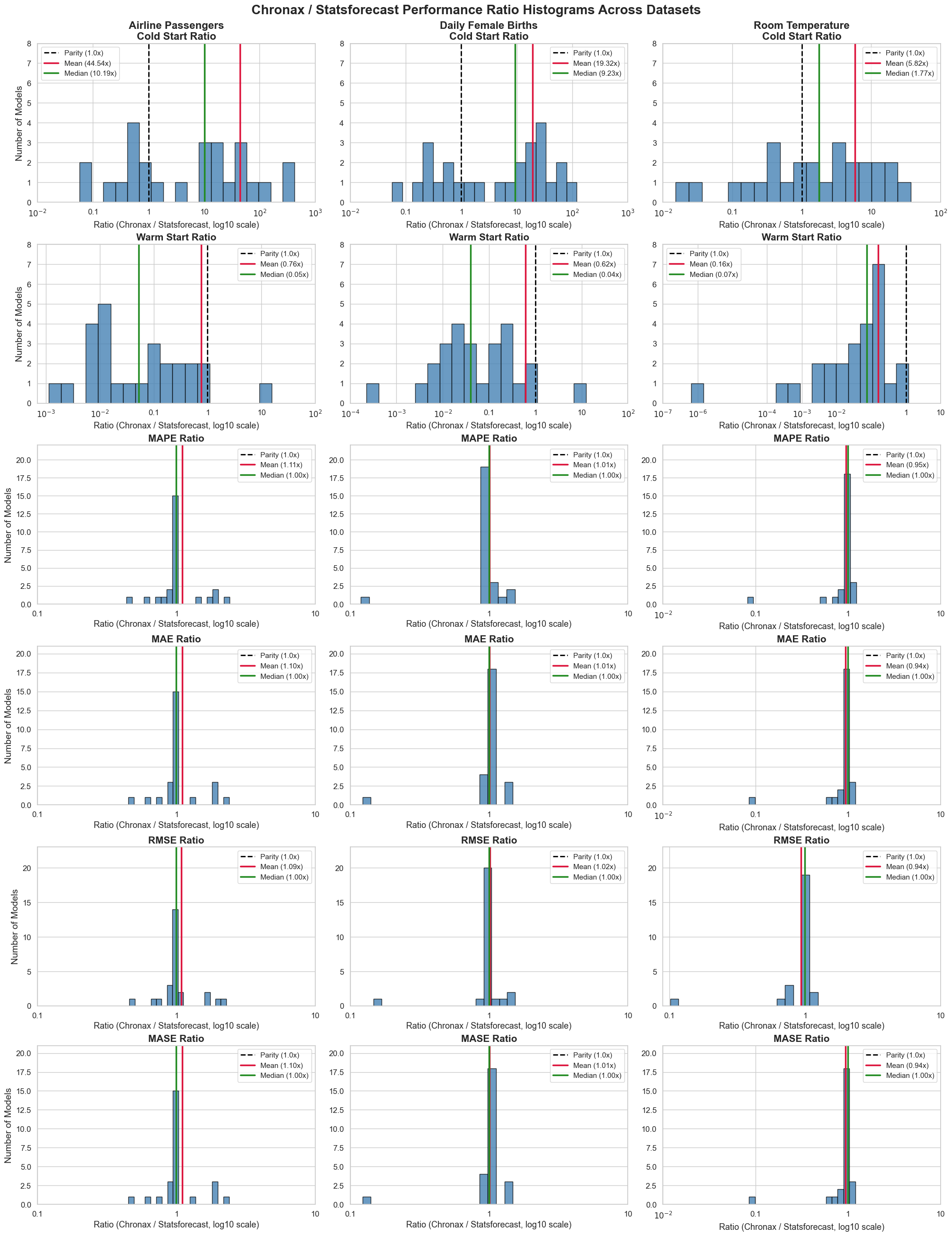}
    \caption{Distribution of Chronax/Statsforecast performance ratios for three datasets and six metrics. Columns show datasets (Airline Passengers, Daily Female Births, Room Temperature), and rows show metrics. Histograms display model counts per logarithmic bin. The dashed line marks parity (1.0$\times$), while red and green lines indicate mean and median. Values below 1.0$\times$ indicate lower latency or error for Chronax. As per Table~\ref{tab:means-medians}, all median and parity lines overlap, with Chronax's mean error metrics higher on the small dataset and lower on the bigger one.}
    \label{fig:combined-ratio-histograms}
\end{figure*}

\subsubsection{Latency and Architectural Trade-offs}

With respect to forecasting accuracy, 
Chronax performs competitively across all three datasets, demonstrating improved performance as dataset size increases, and performing best on the largest evaluated dataset.

For cold start initialization, StatsForecast generally exhibits lower initialization latency. Chronax, by contrast, incurs a higher initialization cost due to JIT compilation, along with lightweight model initialization steps such as parameter setup within the compiled framework. While Chronax’s cold start latency is higher, we do not observe significant variability across models; rather, the initialization cost is consistently higher due to the compilation process itself. 

On the other hand, this upfront cost enables improved performance during warm start execution; once initialized, Chronax consistently demonstrates lower inference latency. With respect to dataset size, Chronax’s cold start initialization remains slower than StatsForecast’s, although the relative difference decreases as dataset size increases (only approximately 6× slower on average for the largest dataset). In contrast, warm start execution is substantially faster, with Chronax requiring only about 16\% of the execution time of StatsForecast on average for the largest dataset. 

These findings reflect a deliberate architectural trade-off common in execution-optimized frameworks that invest more work upfront to improve downstream efficiency.

\section{Conclusion}

This paper introduces Chronax, a JAX-native forecasting library that restructures univariate statistical forecasting and conformal inference around pure functions, composable transformations, and accelerator-ready execution. By representing preprocessing, model evaluation, and multi-horizon prediction as JAX functions transformed via \texttt{jit}, \texttt{vmap}, and \texttt{lax.scan}, Chronax enables scalable multi-series forecasting, end-to-end differentiability, and transparent, hardware-efficient computation. This design allows gradients to propagate through entire sequential pipelines, supporting differentiable hyperparameter tuning and seamless integration with modern machine learning workflows.

Compared to existing libraries such as StatsForecast, HierarchicalForecast, ForeTiS, and Tsururu, Chronax directly addresses three key limitations: limited parallelism over large collections of series, CPU-bound interpreter-centric execution, and object-oriented abstractions that resist differentiable, accelerator-oriented usage. Chronax's XLA-compiled kernels introduce a deliberate architectural trade-off, resulting in slower initialization for certain classical statistical models, but this upfront investment enables more efficient execution during repeated inference calls. During warm execution, Chronax’s functional folds and vectorized evaluation achieve lower inference latency, operating substantially faster than interpreter-bound alternatives. Furthermore, rather than merely preserving predictive accuracy, Chronax achieves lower error rates on a variety of metrics, 
with its accuracy advantage scaling on larger datasets.

Future directions include extending the current univariate, classical model zoo to multivariate and deep learning-based forecasters that share the same functional JAX interface, enabling hybrid pipelines that combine statistical and neural models within a unified framework. Additionally, the conformal inference layer can be further developed to support hierarchical forecasting and improve robustness to distribution shift. Finally, deeper coupling with the broader JAX ecosystem, e.g., differentiable hyperparameter tuning, integration with scientific simulators, and the joint optimization of forecasting components, such as preprocessing filters, latent state representations, and model-agnostic uncertainty wrappers, would position Chronax as a foundation for next-generation forecasting systems.

\bibliographystyle{plainnat}  
\bibliography{references}

\appendix
\newpage
\section{Additional Experimental Results}\label{sec:appendix}

The detailed benchmark results tables in this appendix depict results for three datasets---Airline Passengers, Daily Female Births, and Room Temperature. We report model-wise performance across both Chronax and StatsForecast libraries, including latency metrics (cold start and warm start execution times) and forecasting accuracy metrics (MAPE, MAE, RMSE, and MASE).

\begingroup
\scriptsize %
\setlength{\tabcolsep}{2pt} %
\newcommand{\twolinehead}[2]{\textbf{\begin{tabular}[b]{@{}r@{}}#1\\#2\end{tabular}}}
\begin{longtable}[c]{@{}lllrrrrrrrr@{}} 
\caption{Benchmark Results \\ Rows corresponding to the Chronax framework are highlighted.}
\label{tab:original_results_detailed} \\
\toprule
\textbf{Dataset} & \textbf{Library} & \textbf{Model} & \twolinehead{Series}{length} & \twolinehead{Cold start}{time (s)} & \twolinehead{Warm start}{time (s)} & \twolinehead{Time}{interval (s)} & \textbf{MAPE} & \textbf{MAE} & \textbf{RMSE} & \textbf{MASE} \\ \midrule
\endfirsthead

\multicolumn{11}{c}%
{{\bfseries Table \thetable\ -- continued from previous page}} \\
\toprule
\textbf{Dataset} & \textbf{Library} & \textbf{Model} & \twolinehead{Series}{length} & \twolinehead{Cold start}{time (s)} & \twolinehead{Warm start}{time (s)} & \twolinehead{Time}{interval (s)} & \textbf{MAPE} & \textbf{MAE} & \textbf{RMSE} & \textbf{MASE} \\ \midrule
\endhead

\midrule \multicolumn{11}{r}{{Continued on next page}} \\
\endfoot

\bottomrule
\endlastfoot
\rowcolor{gray!25} Air Passengers & Chronax & ADIDA & 120 & 0.3804 & 0.0004 & 0.0004 & 17.77 & 88.87 & 113.87 & 1.97 \\
\rowcolor{white} Air Passengers & Statsforecast & ADIDA & 120 & 0.6639 & 0.0036 & 0.0055 & 17.81 & 89.04 & 114.04 & 1.97 \\
\rowcolor{gray!25} Air Passengers & Chronax & ARIMA & 120 & 1.7718 & 0.0019 & 0.0019 & 11.45 & 57.68 & 82.13 & 1.28 \\
\rowcolor{white} Air Passengers & Statsforecast & ARIMA & 120 & 0.0170 & 0.0164 & 0.0151 & 18.70 & 93.23 & 118.03 & 2.07 \\
\rowcolor{gray!25} Air Passengers & Chronax & AutoARIMA & 120 & 13.7041 & 0.0172 & 0.0007 & 15.19 & 69.75 & 75.45 & 1.55 \\
\rowcolor{white} Air Passengers & Statsforecast & AutoARIMA & 120 & 0.7421 & 0.7404 & 0.7605 & 10.49 & 51.91 & 72.02 & 1.15 \\
\rowcolor{gray!25} Air Passengers & Chronax & AutoCES & 120 & 1.2726 & 4.17e-05 & 3.35e-05 & 22.70 & 111.36 & 136.18 & 2.47 \\
\rowcolor{white} Air Passengers & Statsforecast & AutoCES & 120 & 22.2269 & 0.0372 & 1.2677 & 12.71 & 61.06 & 70.88 & 1.35 \\
\rowcolor{gray!25} Air Passengers & Chronax & AutoETS & 120 & 12.6266 & 0.0057 & 0.0026 & 12.91 & 59.37 & 66.82 & 1.32 \\
\rowcolor{white} Air Passengers & Statsforecast & AutoETS & 120 & 0.1495 & 0.1484 & 0.1528 & 5.32 & 24.79 & 29.22 & 0.55 \\
\rowcolor{gray!25} Air Passengers & Chronax & AutoMFLES & 120 & 5.9434 & 0.0079 & 0.0075 & 2.97 & 13.71 & 17.72 & 0.30 \\
\rowcolor{white} Air Passengers & Statsforecast & AutoMFLES & 120 & 7.8437 & 0.0729 & 0.1417 & 3.60 & 15.57 & 19.68 & 0.35 \\
\rowcolor{gray!25} Air Passengers & Chronax & AutoTBATS & 120 & 3.0986 & 0.0130 & 0.0151 & 14.15 & 66.61 & 75.56 & 1.48 \\
\rowcolor{white} Air Passengers & Statsforecast & AutoTBATS & 120 & 19.0901 & 4.2171 & 4.1656 & 7.70 & 36.49 & 44.60 & 0.81 \\
\rowcolor{gray!25} Air Passengers & Chronax & AutoTheta & 120 & 5.2890 & 0.0085 & 0.0087 & 18.85 & 93.61 & 117.42 & 2.07 \\
\rowcolor{white} Air Passengers & Statsforecast & AutoTheta & 120 & 0.0125 & 0.0103 & 0.0117 & 18.90 & 93.85 & 117.63 & 2.08 \\
\rowcolor{gray!25} Air Passengers & Chronax & CrostonClassic & 120 & 0.3320 & 4.46e-05 & 3.77e-05 & 16.67 & 83.56 & 108.63 & 1.85 \\
\rowcolor{white} Air Passengers & Statsforecast & CrostonClassic & 120 & 0.4511 & 0.0032 & 0.0043 & 16.67 & 83.56 & 108.63 & 1.85 \\
\rowcolor{gray!25} Air Passengers & Chronax & GARCH & 120 & 6.4052 & 0.0057 & 0.0054 & 44.23 & 206.34 & 219.44 & 4.57 \\
\rowcolor{white} Air Passengers & Statsforecast & GARCH & 120 & 2.0853 & 0.0279 & 0.0267 & 101.00 & 458.24 & 478.31 & 10.15 \\
\rowcolor{gray!25} Air Passengers & Chronax & HistoricAverage & 120 & 0.0876 & 2.14e-05 & 1.34e-05 & 44.23 & 206.34 & 219.44 & 4.57 \\
\rowcolor{white} Air Passengers & Statsforecast & HistoricAverage & 120 & 0.0048 & 0.0031 & 0.0036 & 44.23 & 206.34 & 219.44 & 4.57 \\
\rowcolor{gray!25} Air Passengers & Chronax & Holt & 120 & 0.8623 & 0.0023 & 0.0024 & 17.95 & 89.53 & 113.91 & 1.98 \\
\rowcolor{white} Air Passengers & Statsforecast & Holt & 120 & 0.0150 & 0.0059 & 0.0062 & 18.41 & 91.63 & 115.72 & 2.03 \\
\rowcolor{gray!25} Air Passengers & Chronax & HoltWinters & 120 & 1.8030 & 0.0102 & 0.0100 & 13.06 & 63.72 & 75.90 & 1.41 \\
\rowcolor{white} Air Passengers & Statsforecast & HoltWinters & 120 & 0.0418 & 0.0369 & 0.0365 & 6.90 & 34.46 & 47.01 & 0.76 \\
\rowcolor{gray!25} Air Passengers & Chronax & IMAPA & 120 & 1.0266 & 0.0022 & 0.0021 & 15.86 & 79.70 & 104.95 & 1.77 \\
\rowcolor{white} Air Passengers & Statsforecast & IMAPA & 120 & 0.6133 & 0.0035 & 0.0049 & 17.81 & 89.04 & 114.04 & 1.97 \\
\rowcolor{gray!25} Air Passengers & Chronax & MFLES & 120 & 1.5763 & 0.1064 & 0.0083 & 2.97 & 13.71 & 17.72 & 0.30 \\
\rowcolor{white} Air Passengers & Statsforecast & MFLES & 120 & 2.5072 & 0.0068 & 0.0097 & 4.01 & 18.91 & 23.30 & 0.42 \\
\rowcolor{gray!25} Air Passengers & Chronax & MSTL & 120 & 0.7887 & 0.0012 & 0.0011 & 16.89 & 83.34 & 102.05 & 1.85 \\
\rowcolor{white} Air Passengers & Statsforecast & MSTL & 120 & 0.0235 & 0.0175 & 0.0182 & 16.73 & 80.46 & 93.32 & 1.78 \\
\rowcolor{gray!25} Air Passengers & Chronax & Naive & 120 & 0.1480 & 3.15e-05 & 3.24e-05 & 23.58 & 115.25 & 137.33 & 2.55 \\
\rowcolor{white} Air Passengers & Statsforecast & Naive & 120 & 0.0122 & 0.0031 & 0.0035 & 23.58 & 115.25 & 137.33 & 2.55 \\
\rowcolor{gray!25} Air Passengers & Chronax & RandWalkW/Drift & 120 & 0.0944 & 3.73e-05 & 2.31e-05 & 18.41 & 91.62 & 115.70 & 2.03 \\
\rowcolor{white} Air Passengers & Statsforecast & RandWalkW/Drift & 120 & 0.0052 & 0.0031 & 0.0035 & 18.41 & 91.62 & 115.70 & 2.03 \\
\rowcolor{gray!25} Air Passengers & Chronax & SeasonalExpSmoothing & 120 & 0.2007 & 3.93e-05 & 1.96e-05 & 41.93 & 190.59 & 194.39 & 4.22 \\
\rowcolor{white} Air Passengers & Statsforecast & SeasonalExpSmoothing & 120 & 0.4573 & 0.0032 & 0.0043 & 41.93 & 190.59 & 194.39 & 4.22 \\
\rowcolor{gray!25} Air Passengers & Chronax & SeasonalNaive & 120 & 0.2229 & 0.0007 & 0.0007 & 16.95 & 77.54 & 81.35 & 1.72 \\
\rowcolor{white} Air Passengers & Statsforecast & SeasonalNaive & 120 & 0.0047 & 0.0032 & 0.0037 & 16.95 & 77.54 & 81.35 & 1.72 \\
\rowcolor{gray!25} Air Passengers & Chronax & SeasonalWinAvg & 120 & 0.0524 & 1.92e-05 & 1.62e-05 & 32.52 & 147.99 & 151.28 & 3.28 \\
\rowcolor{white} Air Passengers & Statsforecast & SeasonalWinAvg & 120 & 0.0050 & 0.0031 & 0.0042 & 32.52 & 147.99 & 151.28 & 3.28 \\
\rowcolor{gray!25} Air Passengers & Chronax & SimpleExpSmoothing & 120 & 0.0960 & 2.57e-05 & 1.81e-05 & 17.81 & 89.04 & 114.04 & 1.97 \\
\rowcolor{white} Air Passengers & Statsforecast & SimpleExpSmoothing & 120 & 1.2436 & 0.0032 & 0.0042 & 17.81 & 89.04 & 114.04 & 1.97 \\
\rowcolor{gray!25} Air Passengers & Chronax & TBATS & 120 & 3.3839 & 0.0199 & 0.0204 & 6.83 & 32.62 & 40.46 & 0.72 \\
\rowcolor{white} Air Passengers & Statsforecast & TBATS & 120 & 6.8049 & 0.0537 & 0.0516 & 7.70 & 36.49 & 44.60 & 0.81 \\
\rowcolor{gray!25} Air Passengers & Chronax & TSB & 120 & 0.1212 & 3.98e-05 & 2.12e-05 & 17.81 & 89.04 & 114.04 & 1.97 \\
\rowcolor{white} Air Passengers & Statsforecast & TSB & 120 & 0.4506 & 0.0031 & 0.0045 & 17.81 & 89.04 & 114.04 & 1.97 \\
\rowcolor{gray!25} Air Passengers & Chronax & Theta & 120 & 1.7820 & 0.0036 & 0.0033 & 20.22 & 99.91 & 123.03 & 2.21 \\
\rowcolor{white} Air Passengers & Statsforecast & Theta & 120 & 0.0066 & 0.0052 & 0.0073 & 20.22 & 99.91 & 123.03 & 2.21 \\
\rowcolor{gray!25} Air Passengers & Chronax & WindowAverage & 120 & 0.0483 & 2.05e-05 & 1.30e-05 & 16.46 & 82.55 & 107.65 & 1.83 \\
\rowcolor{white} Air Passengers & Statsforecast & WindowAverage & 120 & 0.0049 & 0.0033 & 0.0079 & 16.46 & 82.55 & 107.65 & 1.83 \\
\midrule
\rowcolor{gray!25} Daily Female Births & Chronax & ADIDA & 341 & 0.3832 & 0.0004 & 0.0004 & 12.08 & 5.44 & 6.64 & 0.86 \\
\rowcolor{white} Daily Female Births & Statsforecast & ADIDA & 341 & 0.6016 & 0.0044 & 0.0049 & 12.07 & 5.43 & 6.62 & 0.86 \\
\rowcolor{gray!25} Daily Female Births & Chronax & ARIMA & 341 & 1.6892 & 0.0019 & 0.0019 & 18.73 & 7.36 & 8.73 & 1.17 \\
\rowcolor{white} Daily Female Births & Statsforecast & ARIMA & 341 & 0.0228 & 0.0201 & 0.0203 & 12.16 & 5.40 & 6.50 & 0.86 \\
\rowcolor{gray!25} Daily Female Births & Chronax & AutoARIMA & 341 & 10.2693 & 0.0139 & 0.0015 & 16.89 & 8.01 & 9.88 & 1.27 \\
\rowcolor{white} Daily Female Births & Statsforecast & AutoARIMA & 341 & 0.3486 & 0.3577 & 0.3462 & 12.31 & 5.40 & 6.41 & 0.86 \\
\rowcolor{gray!25} Daily Female Births & Chronax & AutoCES & 341 & 1.2754 & 3.92e-05 & 3.09e-05 & 12.50 & 5.43 & 6.43 & 0.86 \\
\rowcolor{white} Daily Female Births & Statsforecast & AutoCES & 341 & 22.4988 & 0.1753 & 1.4675 & 13.07 & 5.72 & 6.73 & 0.91 \\
\rowcolor{gray!25} Daily Female Births & Chronax & AutoETS & 341 & 10.3837 & 0.0074 & 0.0021 & 12.17 & 5.38 & 6.43 & 0.85 \\
\rowcolor{white} Daily Female Births & Statsforecast & AutoETS & 341 & 0.3446 & 0.3482 & 0.3427 & 12.29 & 5.38 & 6.37 & 0.85 \\
\rowcolor{gray!25} Daily Female Births & Chronax & AutoMFLES & 341 & 6.1227 & 0.0140 & 0.0140 & 17.05 & 8.18 & 10.45 & 1.30 \\
\rowcolor{white} Daily Female Births & Statsforecast & AutoMFLES & 341 & 5.3699 & 0.1081 & 0.2129 & 13.32 & 6.00 & 7.18 & 0.95 \\
\rowcolor{gray!25} Daily Female Births & Chronax & AutoTBATS & 341 & 2.9946 & 0.0117 & 0.0117 & 13.54 & 5.95 & 6.74 & 0.94 \\
\rowcolor{white} Daily Female Births & Statsforecast & AutoTBATS & 341 & 12.5437 & 2.5173 & 2.5228 & 13.18 & 5.83 & 6.70 & 0.92 \\
\rowcolor{gray!25} Daily Female Births & Chronax & AutoTheta & 341 & 5.3434 & 0.0148 & 0.0147 & 12.06 & 5.41 & 6.58 & 0.86 \\
\rowcolor{white} Daily Female Births & Statsforecast & AutoTheta & 341 & 0.0815 & 0.0705 & 0.0736 & 12.05 & 5.40 & 6.56 & 0.86 \\
\rowcolor{gray!25} Daily Female Births & Chronax & CrostonClassic & 341 & 0.3242 & 0.0001 & 4.55e-05 & 12.07 & 5.43 & 6.62 & 0.86 \\
\rowcolor{white} Daily Female Births & Statsforecast & CrostonClassic & 341 & 0.4434 & 0.0032 & 0.0044 & 12.07 & 5.43 & 6.62 & 0.86 \\
\rowcolor{gray!25} Daily Female Births & Chronax & GARCH & 341 & 7.5769 & 0.0086 & 0.0074 & 12.05 & 5.40 & 6.56 & 0.86 \\
\rowcolor{white} Daily Female Births & Statsforecast & GARCH & 341 & 1.8201 & 0.0267 & 0.0263 & 101.51 & 43.95 & 44.59 & 6.97 \\
\rowcolor{gray!25} Daily Female Births & Chronax & HistoricAverage & 341 & 0.0771 & 0.0001 & 0.0001 & 12.05 & 5.40 & 6.56 & 0.86 \\
\rowcolor{white} Daily Female Births & Statsforecast & HistoricAverage & 341 & 0.0052 & 0.0031 & 0.0035 & 12.05 & 5.40 & 6.56 & 0.86 \\
\rowcolor{gray!25} Daily Female Births & Chronax & Holt & 341 & 0.8256 & 0.0041 & 0.0045 & 12.24 & 5.68 & 7.27 & 0.90 \\
\rowcolor{white} Daily Female Births & Statsforecast & Holt & 341 & 0.0252 & 0.0236 & 0.0239 & 12.31 & 5.37 & 6.36 & 0.85 \\
\rowcolor{gray!25} Daily Female Births & Chronax & HoltWinters & 341 & 1.8571 & 0.0667 & 0.0683 & 13.35 & 5.83 & 6.80 & 0.92 \\
\rowcolor{white} Daily Female Births & Statsforecast & HoltWinters & 341 & 0.0818 & 0.0810 & 0.0834 & 12.96 & 5.67 & 6.71 & 0.90 \\
\rowcolor{gray!25} Daily Female Births & Chronax & IMAPA & 341 & 1.0295 & 0.0024 & 0.0022 & 12.12 & 5.38 & 6.46 & 0.85 \\
\rowcolor{white} Daily Female Births & Statsforecast & IMAPA & 341 & 0.5983 & 0.0037 & 0.0059 & 12.07 & 5.43 & 6.62 & 0.86 \\
\rowcolor{gray!25} Daily Female Births & Chronax & MFLES & 341 & 1.4365 & 0.1070 & 0.0137 & 17.05 & 8.18 & 10.45 & 1.30 \\
\rowcolor{white} Daily Female Births & Statsforecast & MFLES & 341 & 5.3658 & 0.0085 & 0.0128 & 18.26 & 8.68 & 10.80 & 1.38 \\
\rowcolor{gray!25} Daily Female Births & Chronax & MSTL & 341 & 0.8115 & 0.0012 & 0.0011 & 14.11 & 6.15 & 7.14 & 0.97 \\
\rowcolor{white} Daily Female Births & Statsforecast & MSTL & 341 & 0.0492 & 0.0449 & 0.0483 & 14.38 & 6.37 & 7.45 & 1.01 \\
\rowcolor{gray!25} Daily Female Births & Chronax & Naive & 341 & 0.0741 & 0.0001 & 4.48e-05 & 16.95 & 8.04 & 9.91 & 1.28 \\
\rowcolor{white} Daily Female Births & Statsforecast & Naive & 341 & 0.0064 & 0.0044 & 0.0035 & 16.95 & 8.04 & 9.91 & 1.28 \\
\rowcolor{gray!25} Daily Female Births & Chronax & RandWalkW/Drift & 341 & 0.0850 & 3.84e-05 & 2.55e-05 & 16.89 & 8.01 & 9.87 & 1.27 \\
\rowcolor{white} Daily Female Births & Statsforecast & RandWalkW/Drift & 341 & 0.0054 & 0.0032 & 0.0035 & 16.89 & 8.01 & 9.87 & 1.27 \\
\rowcolor{gray!25} Daily Female Births & Chronax & SeasonalExpSmoothing & 341 & 0.1651 & 0.0001 & 0.0001 & 15.27 & 6.62 & 7.63 & 1.05 \\
\rowcolor{white} Daily Female Births & Statsforecast & SeasonalExpSmoothing & 341 & 0.4422 & 0.0033 & 0.0043 & 15.27 & 6.62 & 7.63 & 1.05 \\
\rowcolor{gray!25} Daily Female Births & Chronax & SeasonalNaive & 341 & 0.1947 & 0.0007 & 0.0006 & 18.04 & 7.92 & 9.87 & 1.26 \\
\rowcolor{white} Daily Female Births & Statsforecast & SeasonalNaive & 341 & 0.0049 & 0.0031 & 0.0035 & 18.04 & 7.92 & 9.87 & 1.26 \\
\rowcolor{gray!25} Daily Female Births & Chronax & SeasonalWinAvg & 341 & 0.0496 & 2.00e-05 & 1.50e-05 & 17.12 & 7.39 & 8.31 & 1.17 \\
\rowcolor{white} Daily Female Births & Statsforecast & SeasonalWinAvg & 341 & 0.0050 & 0.0031 & 0.0043 & 17.12 & 7.39 & 8.31 & 1.17 \\
\rowcolor{gray!25} Daily Female Births & Chronax & SimpleExpSmoothing & 341 & 0.0797 & 2.78e-05 & 2.28e-05 & 14.73 & 7.00 & 8.85 & 1.11 \\
\rowcolor{white} Daily Female Births & Statsforecast & SimpleExpSmoothing & 341 & 0.4390 & 0.0031 & 0.0042 & 14.73 & 7.00 & 8.85 & 1.11 \\
\rowcolor{gray!25} Daily Female Births & Chronax & TBATS & 341 & 3.2495 & 0.0171 & 0.0170 & 12.63 & 5.43 & 6.29 & 0.86 \\
\rowcolor{white} Daily Female Births & Statsforecast & TBATS & 341 & 6.7732 & 0.0959 & 0.0967 & 12.28 & 5.69 & 7.20 & 0.90 \\
\rowcolor{gray!25} Daily Female Births & Chronax & TSB & 341 & 0.1098 & 3.14e-05 & 2.24e-05 & 14.73 & 7.00 & 8.85 & 1.11 \\
\rowcolor{white} Daily Female Births & Statsforecast & TSB & 341 & 0.4711 & 0.0032 & 0.0042 & 14.73 & 7.00 & 8.85 & 1.11 \\
\rowcolor{gray!25} Daily Female Births & Chronax & Theta & 341 & 1.7261 & 0.0047 & 0.0044 & 12.05 & 5.39 & 6.54 & 0.86 \\
\rowcolor{white} Daily Female Births & Statsforecast & Theta & 341 & 0.0144 & 0.0124 & 0.0138 & 12.05 & 5.39 & 6.54 & 0.86 \\
\rowcolor{gray!25} Daily Female Births & Chronax & WindowAverage & 341 & 0.0423 & 2.07e-05 & 1.27e-05 & 12.31 & 5.38 & 6.36 & 0.85 \\
\rowcolor{white} Daily Female Births & Statsforecast & WindowAverage & 341 & 0.0049 & 0.0032 & 0.0043 & 12.31 & 5.38 & 6.36 & 0.85 \\
\midrule
\rowcolor{gray!25} Room Temperatures & Chronax & ADIDA & 7032 & 0.3859 & 0.0008 & 0.0008 & 6.87 & 1.59 & 1.83 & 3.55 \\
\rowcolor{white} Room Temperatures & Statsforecast & ADIDA & 7032 & 0.6123 & 0.0043 & 0.0067 & 6.86 & 1.59 & 1.83 & 3.54 \\
\rowcolor{gray!25} Room Temperatures & Chronax & ARIMA & 7032 & 1.7198 & 0.0140 & 0.0134 & 5.57 & 1.34 & 1.59 & 2.99 \\
\rowcolor{white} Room Temperatures & Statsforecast & ARIMA & 7032 & 0.0799 & 0.0810 & 0.0759 & 5.58 & 1.35 & 1.59 & 2.99 \\
\rowcolor{gray!25} Room Temperatures & Chronax & AutoARIMA & 7032 & 13.3224 & 0.0159 & 0.0045 & 5.60 & 1.35 & 1.59 & 3.00 \\
\rowcolor{white} Room Temperatures & Statsforecast & AutoARIMA & 7032 & 4.7570 & 4.7766 & 4.7759 & 5.71 & 1.34 & 1.57 & 2.98 \\
\rowcolor{gray!25} Room Temperatures & Chronax & AutoCES & 7032 & 1.4025 & 4.25e-05 & 3.08e-05 & 9.57 & 2.20 & 2.62 & 4.91 \\
\rowcolor{white} Room Temperatures & Statsforecast & AutoCES & 7032 & 91.4692 & 64.4936 & 67.6398 & 10.64 & 2.67 & 3.63 & 5.95 \\
\rowcolor{gray!25} Room Temperatures & Chronax & AutoETS & 7032 & 23.0873 & 0.0058 & 0.0023 & 4.66 & 1.16 & 1.53 & 2.58 \\
\rowcolor{white} Room Temperatures & Statsforecast & AutoETS & 7032 & 8.1513 & 8.0565 & 8.0164 & 5.61 & 1.35 & 1.59 & 3.00 \\
\rowcolor{gray!25} Room Temperatures & Chronax & AutoMFLES & 7032 & 8.1671 & 0.1241 & 0.1334 & 4.70 & 1.15 & 1.42 & 2.56 \\
\rowcolor{white} Room Temperatures & Statsforecast & AutoMFLES & 7032 & 65.9265 & 58.6217 & 118.1957 & 4.78 & 1.17 & 1.46 & 2.60 \\
\rowcolor{gray!25} Room Temperatures & Chronax & AutoTBATS & 7032 & 3.0661 & 0.0288 & 0.0289 & 4.87 & 1.18 & 1.47 & 2.62 \\
\rowcolor{white} Room Temperatures & Statsforecast & AutoTBATS & 7032 & 95.4781 & 83.5333 & 83.3852 & 4.10 & 0.98 & 1.18 & 2.18 \\
\rowcolor{gray!25} Room Temperatures & Chronax & AutoTheta & 7032 & 6.3909 & 0.3354 & 0.3329 & 5.45 & 1.36 & 1.79 & 3.02 \\
\rowcolor{white} Room Temperatures & Statsforecast & AutoTheta & 7032 & 4.2738 & 4.3127 & 4.9150 & 5.45 & 1.36 & 1.79 & 3.02 \\
\rowcolor{gray!25} Room Temperatures & Chronax & CrostonClassic & 7032 & 0.3828 & 0.0004 & 0.0004 & 7.22 & 1.67 & 1.92 & 3.71 \\
\rowcolor{white} Room Temperatures & Statsforecast & CrostonClassic & 7032 & 0.4451 & 0.0037 & 0.0056 & 7.22 & 1.67 & 1.92 & 3.71 \\
\rowcolor{gray!25} Room Temperatures & Chronax & GARCH & 7032 & 11.2155 & 0.0384 & 0.0380 & 8.28 & 2.07 & 2.58 & 4.60 \\
\rowcolor{white} Room Temperatures & Statsforecast & GARCH & 7032 & 1.8963 & 0.0782 & 0.0843 & 101.33 & 24.24 & 25.39 & 53.94 \\
\rowcolor{gray!25} Room Temperatures & Chronax & HistoricAverage & 7032 & 0.0902 & 0.0001 & 0.0001 & 8.28 & 2.07 & 2.58 & 4.60 \\
\rowcolor{white} Room Temperatures & Statsforecast & HistoricAverage & 7032 & 0.0054 & 0.0035 & 0.0039 & 8.28 & 2.07 & 2.58 & 4.60 \\
\rowcolor{gray!25} Room Temperatures & Chronax & Holt & 7032 & 0.8942 & 0.0242 & 0.0247 & 5.65 & 1.36 & 1.61 & 3.03 \\
\rowcolor{white} Room Temperatures & Statsforecast & Holt & 7032 & 0.1756 & 0.1762 & 0.1787 & 5.61 & 1.35 & 1.59 & 3.00 \\
\rowcolor{gray!25} Room Temperatures & Chronax & HoltWinters & 7032 & 3.1843 & 1.2879 & 1.2806 & 4.05 & 1.01 & 1.35 & 2.24 \\
\rowcolor{white} Room Temperatures & Statsforecast & HoltWinters & 7032 & 1.7016 & 1.6955 & 1.6940 & 7.35 & 1.71 & 1.99 & 3.80 \\
\rowcolor{gray!25} Room Temperatures & Chronax & IMAPA & 7032 & 1.0110 & 0.0050 & 0.0048 & 7.74 & 1.78 & 2.08 & 3.96 \\
\rowcolor{white} Room Temperatures & Statsforecast & IMAPA & 7032 & 0.6071 & 0.0042 & 0.0067 & 6.86 & 1.59 & 1.83 & 3.54 \\
\rowcolor{gray!25} Room Temperatures & Chronax & MFLES & 7032 & 1.5401 & 0.1300 & 0.1176 & 4.70 & 1.15 & 1.42 & 2.56 \\
\rowcolor{white} Room Temperatures & Statsforecast & MFLES & 7032 & 4.4536 & 1.8329 & 3.6509 & 6.14 & 1.52 & 1.99 & 3.39 \\
\rowcolor{gray!25} Room Temperatures & Chronax & MSTL & 7032 & 0.7934 & 0.0624 & 0.0561 & 3.97 & 0.92 & 1.09 & 2.05 \\
\rowcolor{white} Room Temperatures & Statsforecast & MSTL & 7032 & 0.7241 & 0.7216 & 0.7126 & 4.10 & 1.00 & 1.34 & 2.23 \\
\rowcolor{gray!25} Room Temperatures & Chronax & Naive & 7032 & 0.0794 & 0.0001 & 0.0001 & 5.61 & 1.35 & 1.59 & 3.00 \\
\rowcolor{white} Room Temperatures & Statsforecast & Naive & 7032 & 0.0054 & 0.0035 & 0.0040 & 5.61 & 1.35 & 1.59 & 3.00 \\
\rowcolor{gray!25} Room Temperatures & Chronax & RandWalkW/Drift & 7032 & 0.0849 & 0.0001 & 0.0001 & 5.60 & 1.35 & 1.59 & 3.00 \\
\rowcolor{white} Room Temperatures & Statsforecast & RandWalkW/Drift & 7032 & 0.0056 & 0.0036 & 0.0040 & 5.60 & 1.35 & 1.59 & 3.00 \\
\rowcolor{gray!25} Room Temperatures & Chronax & SeasonalExpSmoothing & 7032 & 0.1833 & 0.0005 & 0.0005 & 3.77 & 0.88 & 1.01 & 1.97 \\
\rowcolor{white} Room Temperatures & Statsforecast & SeasonalExpSmoothing & 7032 & 0.4584 & 0.0036 & 0.0054 & 3.77 & 0.88 & 1.01 & 1.97 \\
\rowcolor{gray!25} Room Temperatures & Chronax & SeasonalNaive & 7032 & 0.1925 & 0.0007 & 0.0006 & 1.98 & 0.47 & 0.57 & 1.06 \\
\rowcolor{white} Room Temperatures & Statsforecast & SeasonalNaive & 7032 & 0.0052 & 0.0036 & 0.0040 & 1.98 & 0.47 & 0.57 & 1.06 \\
\rowcolor{gray!25} Room Temperatures & Chronax & SeasonalWinAvg & 7032 & 0.0476 & 1.93e-05 & 1.59e-05 & 4.17 & 0.98 & 1.49 & 2.18 \\
\rowcolor{white} Room Temperatures & Statsforecast & SeasonalWinAvg & 7032 & 0.0053 & 0.0035 & 0.0052 & 4.17 & 0.98 & 1.49 & 2.18 \\
\rowcolor{gray!25} Room Temperatures & Chronax & SimpleExpSmoothing & 7032 & 0.0776 & 0.0001 & 0.0001 & 6.86 & 1.59 & 1.83 & 3.54 \\
\rowcolor{white} Room Temperatures & Statsforecast & SimpleExpSmoothing & 7032 & 0.4450 & 0.0035 & 0.0052 & 6.86 & 1.59 & 1.83 & 3.54 \\
\rowcolor{gray!25} Room Temperatures & Chronax & TBATS & 7032 & 3.6609 & 0.3444 & 0.3305 & 7.29 & 1.68 & 1.95 & 3.75 \\
\rowcolor{white} Room Temperatures & Statsforecast & TBATS & 7032 & 8.9820 & 2.1793 & 2.2344 & 5.93 & 1.47 & 1.94 & 3.28 \\
\rowcolor{gray!25} Room Temperatures & Chronax & TSB & 7032 & 0.1219 & 0.0001 & 0.0001 & 6.86 & 1.59 & 1.83 & 3.54 \\
\rowcolor{white} Room Temperatures & Statsforecast & TSB & 7032 & 0.4479 & 0.0036 & 0.0052 & 6.86 & 1.59 & 1.83 & 3.54 \\
\rowcolor{gray!25} Room Temperatures & Chronax & Theta & 7032 & 2.5806 & 0.1693 & 0.1620 & 5.45 & 1.36 & 1.79 & 3.02 \\
\rowcolor{white} Room Temperatures & Statsforecast & Theta & 7032 & 0.8926 & 0.8812 & 0.8816 & 5.45 & 1.36 & 1.79 & 3.02 \\
\rowcolor{gray!25} Room Temperatures & Chronax & WindowAverage & 7032 & 0.0440 & 2.02e-05 & 1.23e-05 & 6.21 & 1.46 & 1.65 & 3.25 \\
\rowcolor{white} Room Temperatures & Statsforecast & WindowAverage & 7032 & 0.0054 & 0.0035 & 0.0050 & 6.21 & 1.46 & 1.65 & 3.25 \\
\end{longtable}
\endgroup

\newpage

\begin{figure}[p]
    \centering
    \includegraphics[width=\textwidth]{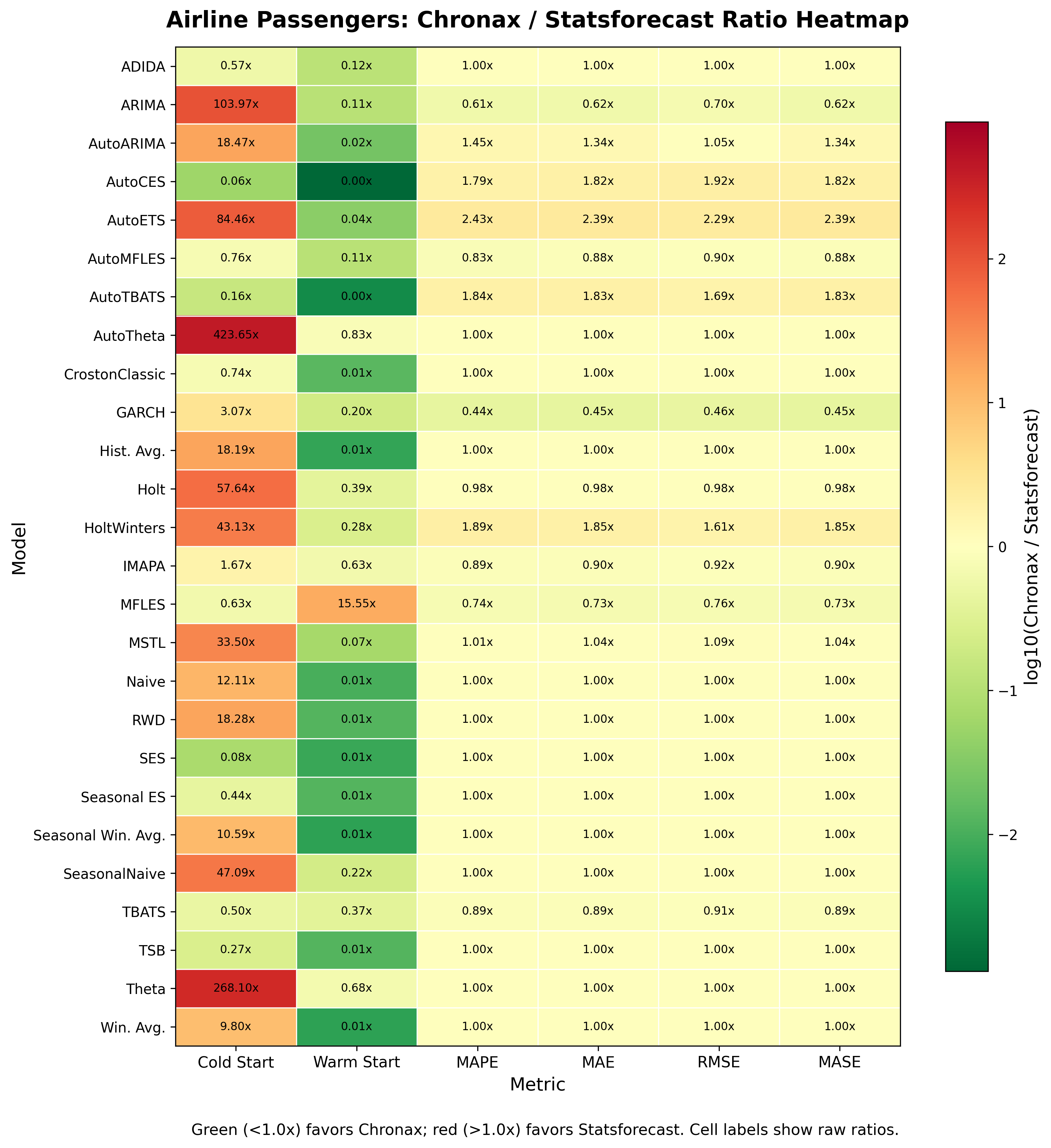}
    \caption{Heatmap of Chronax/Statsforecast benchmarking ratios for the Airline Passengers dataset across six metrics. Rows correspond to forecasting models and columns correspond to Cold Start, Warm Start, MAPE, MAE, RMSE, and MASE. Cell labels report raw ratios, while color encodes $\log_{10}(\mathrm{Chronax}/\mathrm{Statsforecast})$. Green values ($<1.0\times$) favor Chronax and red values ($>1.0\times$) favor Statsforecast.}
    \label{fig:airline-ratio-heatmap}
\end{figure}

\clearpage

\begin{figure}[p]
    \centering
    \includegraphics[width=\textwidth]{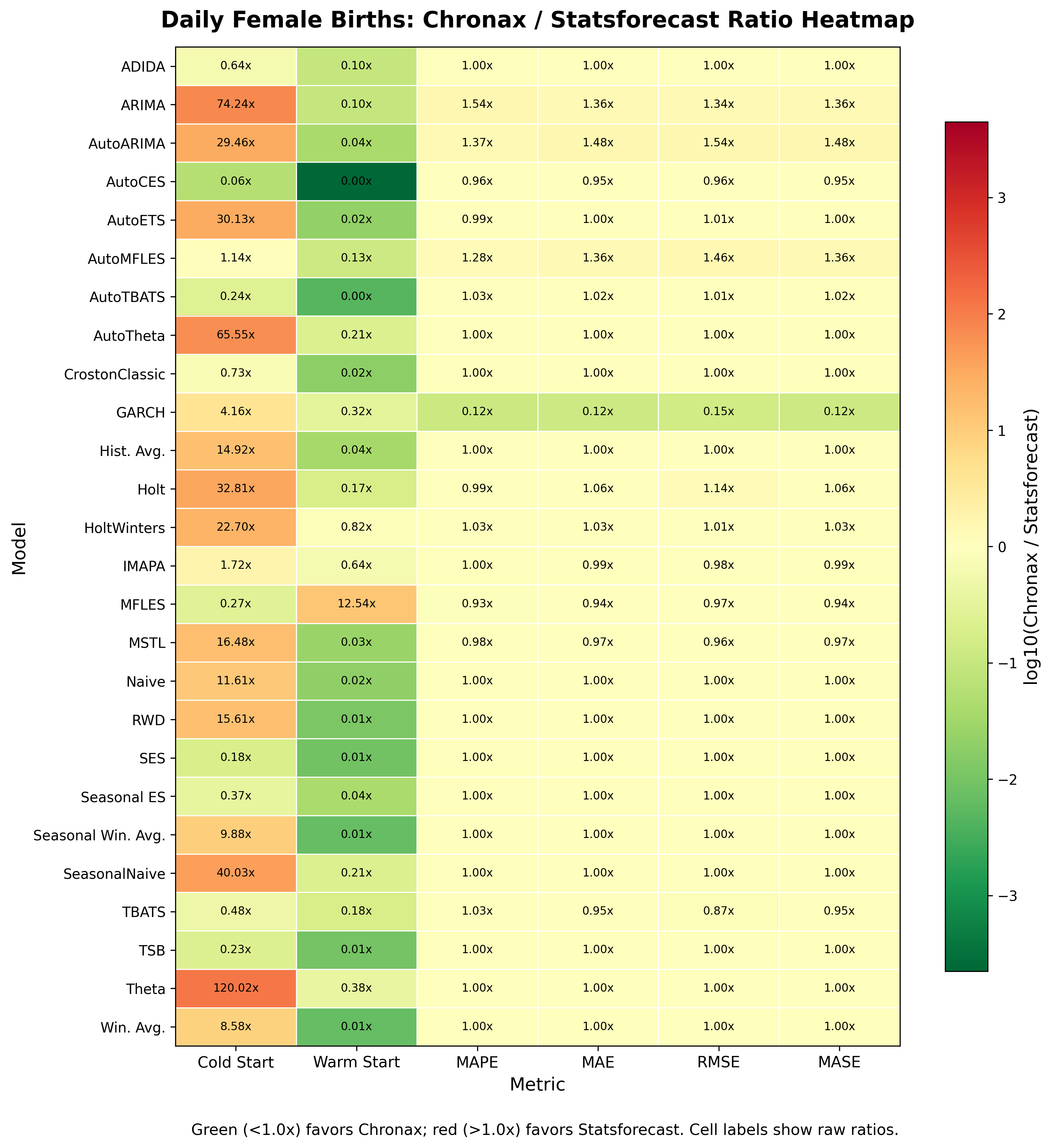}
    \caption{Heatmap of Chronax/Statsforecast benchmarking ratios for the Daily Female Births dataset across six metrics. Rows correspond to forecasting models and columns correspond to Cold Start, Warm Start, MAPE, MAE, RMSE, and MASE. Cell labels report raw ratios, while color encodes $\log_{10}(\mathrm{Chronax}/\mathrm{Statsforecast})$. Green values ($<1.0\times$) favor Chronax and red values ($>1.0\times$) favor Statsforecast.}
    \label{fig:births-ratio-heatmap}
\end{figure}

\clearpage

\begin{figure}[p]
    \centering
    \includegraphics[width=\textwidth]{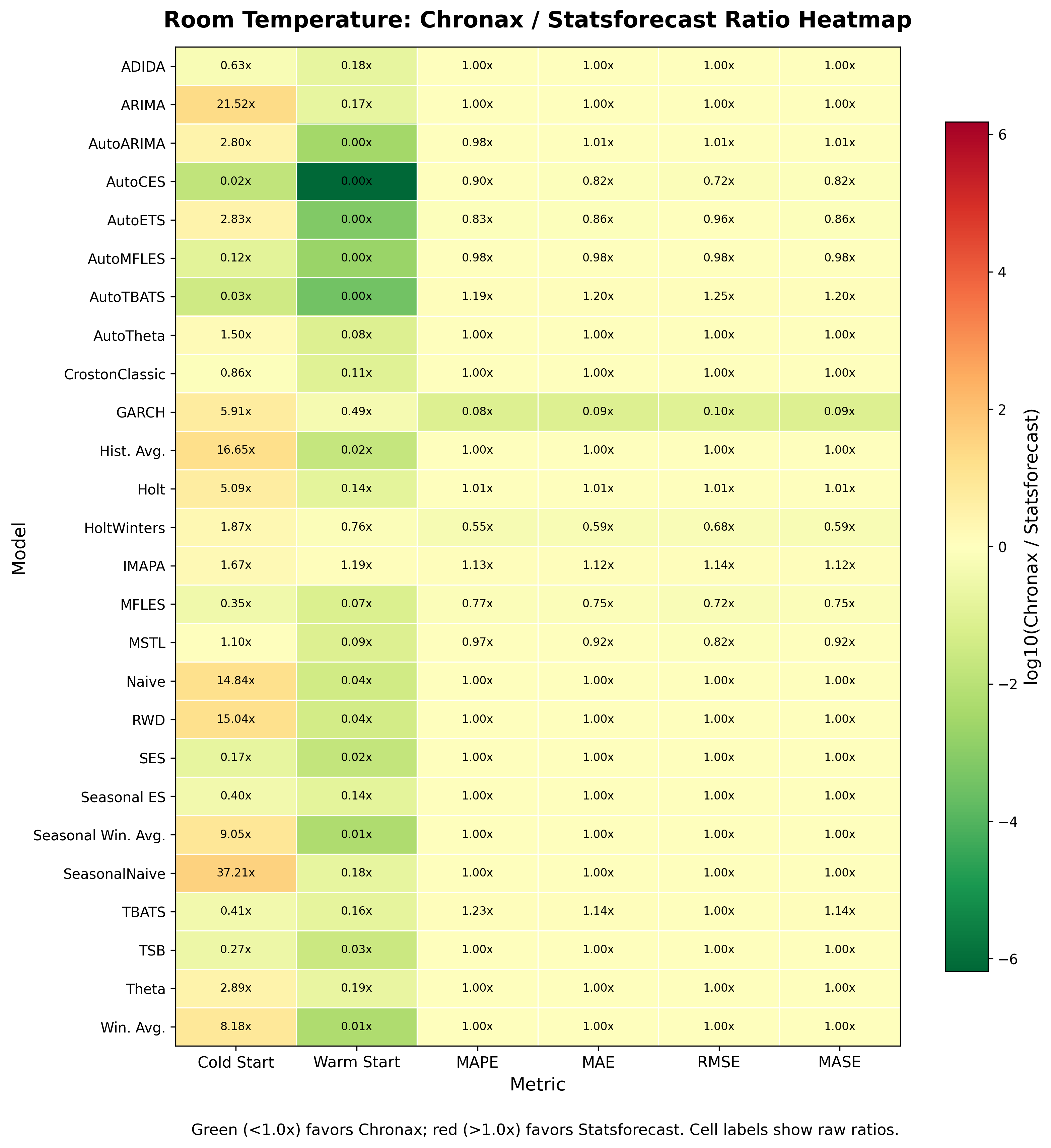}
    \caption{Heatmap of Chronax/Statsforecast benchmarking ratios for the Room Temperature dataset across six metrics. Rows correspond to forecasting models and columns correspond to Cold Start, Warm Start, MAPE, MAE, RMSE, and MASE. Cell labels report raw ratios, while color encodes $\log_{10}(\mathrm{Chronax}/\mathrm{Statsforecast})$. Green values ($<1.0\times$) favor Chronax and red values ($>1.0\times$) favor Statsforecast.}
    \label{fig:room-ratio-heatmap}
\end{figure}

\aboutsimulacrum
\end{document}